\documentclass[11pt,a4paper]{article}
\usepackage[usenames,dvipsnames]{color}
\usepackage{multirow}
\usepackage[hyperref]{mmtm}
\usepackage{times}
\usepackage{latexsym}
\usepackage{graphicx}
\usepackage{amsmath}

\usepackage{microtype}

\aclfinalcopy % Uncomment this line for the final submission
\title{
MMTM: Multi-Tasking Multi-Decoder Transformer for Math Word Problems
}

\author{Keyur Faldu \thanks{\hspace{0.1cm}Correspondence to k@embibe.com}\\
  Embibe \\\And
  Amit Sheth \\
  University of\\ South Carolina \\\And
  Prashant Kikani \thanks{\hspace{0.1cm}Owned implementation and experiments.}\\
  Embibe \\\And
  Darshan Patel \\
  Embibe \\
  }

\date{}

\begin{document}
\maketitle
\begin{abstract}

\end{abstract}
Recently, quite a few novel neural architectures were derived to solve math word problems by predicting expression trees. These architectures varied from seq2seq models, including encoders leveraging graph relationships combined with tree decoders. These models achieve good performance on various MWPs datasets but perform poorly when applied to an adversarial challenge dataset, SVAMP. We present a novel model MMTM that leverages multi-tasking and multi-decoder during pre-training. It creates variant tasks by deriving labels using pre-order, in-order and post-order traversal of expression trees, and uses task-specific decoders in a multi-tasking framework. We leverage transformer architectures with lower dimensionality and initialize weights from RoBERTa model. MMTM model achieves better mathematical reasoning ability and generalisability, which we demonstrate by outperforming the best state of the art baseline models from Seq2Seq, GTS, and Graph2Tree with a relative improvement of 19.4\% on an adversarial challenge dataset SVAMP. 

\section{Introduction}

\begin{figure}[h]
    \centering
    \includegraphics[scale=0.6]{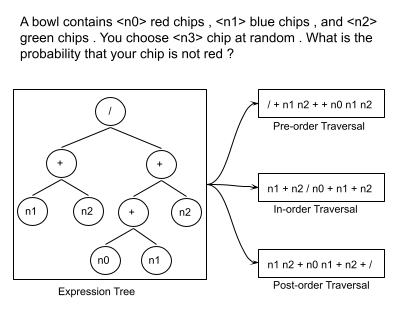}
    \caption{Sample Math Word Problem and Traversal Sequences of its Expression Trees}
    \label{fig:math_reason}
\end{figure}

The field of natural language processing is fundamentally impacted by novel deep learning techniques, like transformer-based language models (TLMs) \cite{devlin2018bert} \cite{raffel2019exploring}. There is a flurry of research and analysis on the natural language understanding ability of such TLMs \cite{wang2018glue} \cite{wang2019superglue} \cite{faldu2021ki}. NLP applications like comprehension, open-book question answering, and conversational chat-bots are now getting adopted for various specific use-cases. Mathematical reasoning is one important aspect of human intelligence, and many times natural language understanding would need to encompass the mathematical ability to understand numerical quantities and manipulate them to deal with the situation. For example, comprehension based question answering tasks (DROP) involves operations like counting, comparison, addition and subtraction to answer the question \cite{dua2019drop}. 

There has been a long research history with a recent surge in solving math word problems \cite{faldu2021towards}. A math problem narrated using natural language is called a math word problem. Math word problems would narrate quantities with natural language descriptions, and a question about an unknown quantity or set of unknown quantities which could be derived by applying mathematical laws, commonsense world knowledge and explained quantities. The ability to solve math word problems could be very useful to the education domain in applications like improving learning outcomes \cite{faldu2020adaptive} and generating diagnostic tests \cite{dhavala2020auto} etc.  The efforts to solve math word problems using the software have been sparingly made since the 1960s \cite{feigenbaum1963computers} \cite{bobrow1964natural}. Rule-based and pattern matching approaches have been explored to select a set of relevant mathematical rules and apply them to arrive at an answer \cite{briars1984integrated} \cite{fletcher1985understanding} \cite{dellarosa1986computer} \cite{bakman2007robust} \cite{yuhui2010frame}. Semantic parsing based approach process input narratives to extract quantities and their semantic relations to each other, like entities, attributes, values and how entities are associated with each other. Afterwards, it process the quantities extracted in semantic structure using mathematical laws to compute the answer \cite{liguda2012modeling} \cite{shi2015automatically} \cite{koncel2015parsing} \cite{huang2017learning}. Statistical and machine learning approaches to solve MWPs leverages machine learning techniques like SVM, Probabilistic models, Bayesian models etc are used to extract semantic structures and classify the sub-type of mathematical problem and search the relevant mathematical law from all candidates \cite{amnueypornsakul2014machine} \cite{hosseini2014learning} \cite{zhou2015learn} \cite{mitra2016learning} \cite{roy2018mapping}. The recent success of deep learning and transfer learning also has proven very effective in solving MWPs \cite{wang2017deep} \cite{liang2021mwp}.

Generally, neural models aim to generate expression trees for MWPs, which are evaluated to compute the final answer. Such expression trees also give better interpretability. Encoder-decoder based neural approaches which are graph-based encoders and tree decoders are among the state of the art model performing well on MWPs datasets.  However adversarial analysis of well-performing neural approaches reveals that their generalization and reasoning ability is questionable even on simpler math word problems, specifically for the smaller datasets \cite{patel2021nlp} \cite{miao2021diverse}. 

On the other hand, Transformer models on very large synthetic datasets have achieved near-perfect accuracy on complex problems like calculus and integration, which demonstrates that larger datasets help the model to perform better to solve MWPs \cite{saxton2019analysing} \cite{lample2019deep}.

We present a novel approach, which triples the size of MWPs training corpus and learns semantics with better generalizability on the adversarial datasets. 

We propose a novel architecture and training method Multi-tasking Multi-decoder Transformer for Math Word Problems (MMTM) which creates a three times larger training corpus and aims to learn semantic relationships generically. It significantly improves performance on the SVAMP challenge. Which comprises the following novelties.

1) We propose a data augmentation technique that triples the training corpus size by leveraging pre-order, post-order and in-order traversal of expression trees. We create three traversal-specific tasks and datasets from an input MWP dataset.

2) We propose a shared encoder with traversal order specific decoders and a shared encoder. Such a setting guides the decoder to learn traversal sequencing and the shared encoder to learn semantic relationships generically.   

3) MMTM model is pre-trained using multi-tasking over all the traversal specific variant datasets, and fine-tuned on the actual MWP dataset with expression trees represented using pre-order traversal.

4) We analyse various configuration settings for transformer architecture. A model with a single layer, 64 dimensions for hidden representation, and initializing vocabulary embeddings with principal components from pre-trained RoBERTa generalize well on the adversarial SVAMP challenge.

We plan to release traversal specific datasets for the SVAMP challenge and also the code-base for Multi-Tasking Multi-Decoder Transformer for Math Word Problems to collaborate with the research community effectively.

\section{Related Work}

Statistical and machine learning techniques could not scale well on the larger and more diverse datasets. The sub-linear gain in performance with an increase in training data size suggests the need of developing new approaches \cite{huang2016well}. 

In recent years, quite a few neural approaches have been explored to solve math word problems. A class of neural approaches attempt to predict expression trees, which are evaluated to compute the final answer \cite{wang2017deep}. They have been broadly classified into sequence-to-sequence, transformer-to-tree, sequence-to-tree and graph-to-tree based encoder-decoder architectures. Sequence-to-sequence approaches use GRU and LSTM architectures for their encoder and decoder. Tree decoders attempt to leverage the semantic relationships between operators and quantities and their groups using the parent-child relationships of trees. MWP-BERT is an example of a transformer-to-tree approach where there is a tree decoder on the pre-trained BERT on MWP corpus \cite{liang2021mwp}. Goal-driven tree structure model leverages GRU based encoder and tree decoder to predict expression trees \cite{xie2019goal}. On the other hand, a graph of entities, quantities, and their relationships could also help the encoder learn the semantic structure using GNNs and GCN for better performance in predicting expression trees \cite{zhang2020graph} \cite{ran2019numnet}. These techniques have achieved significant performance on datasets like Alg54, MAWPS. MathQA, Math23K, etc \cite{koncel2016mawps} \cite{kushman2014learning} \cite{amini2019mathqa} \cite{wang2017deep}.

The diversity of such datasets and the reasoning ability of the models came to question when researchers analysed datasets like MathQA have very low lexicon diversity scores, which could have overlapping templates between train and test set, which could lead to overfitting of the model. A challenge dataset SVAMP was created by adding 9 adversarial variants of a question, comprising a total of 1000 questions \cite{patel2021nlp}. GTS and Graph-to-tree models could not generalise well enough on this SVAMP dataset highlighting the lack of mathematical reasoning ability of such models, especially when the training corpus is small \cite{xie2019goal}, \cite{zhang2020graph}. Graph relationships in such models are heuristic-based, a more robust way to learn such relationships needs to be developed. Multi-tasking pre-training of MMTM model on traversal-specific datasets guides shared encoder to learn semantic relations in a generic way and task-specific decoder to learn traversal sequencing. 

Transformer models have been successfully trained on very large datasets \cite{saxton2019analysing} \cite{lample2019deep}, but on the smaller datasets, they could not outperform SOTA models like GTS and Graph2Tree. Our novel approach demonstrates a strategy to leverage transformer models to outperform GTS and Graph2Tree models even on smaller datasets. It triples the training corpus size by leveraging pre-order, in-order and post-order traversal of expression trees.

\begin{figure*}[h]
    \centering
    \includegraphics[scale=0.5]{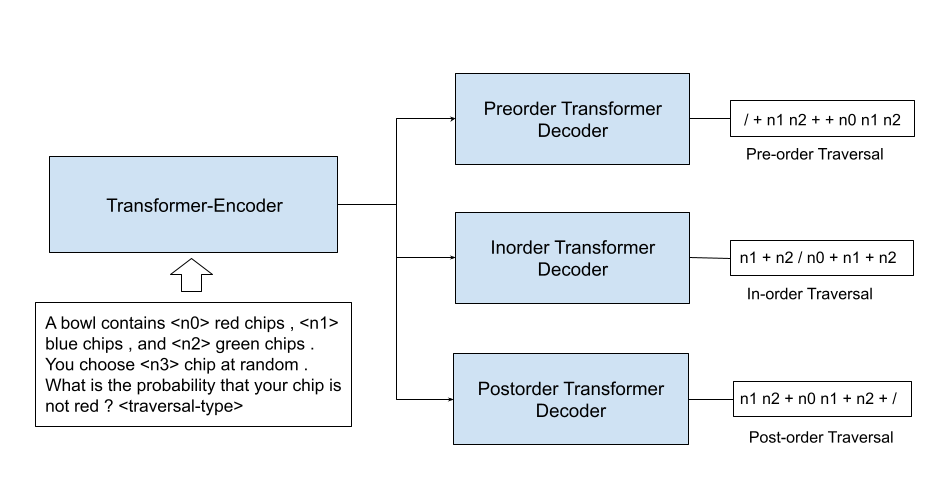}
    \caption{Architecture Diagram for Multi-Tasking Multi-Decoder Transformer Model}
    \label{fig:math_reason}
\end{figure*}

\section{Multi Tasking Multi Decoder Transformer}

We propose a novel multi-tasking multi-decoder transformer model for solving math word problems. We formulate a problem to generate expression trees, which are evaluated afterwards to compute the final answer for a one-unknown arithmetic problem. Our model has two important considerations, it creates three variant tasks to enable multi-tasking settings, and it uses a shared encoder and task-specific decoder, which would help the shared encoder learn semantics useful for each task-specific decoder. We explain in detail (1) how we generate variant tasks, (2) the motivation and architecture of shared encoder and multi decoder, and (3) pre-training training leveraging multi-tasking settings for multi-decoder architecture in the following sections. 

\subsection{Traversal Specific Tasks}
The labels for the input math word problems are expression trees. A transformer-based decoder could generate the sequence of tokens, we need to transform expression trees into sequential order. The most popular and effective way to transform the expression tree in a sequence is using pre-order traversal. Where an operator is chosen first, operands are chosen later, and operands could either be a numerical quantity or an operator itself. If a selected operand is an operator, then the process is repeated recursively. 

We propose to generate other sequences from the expression trees, based on in-order traversal and pos-order traversal. So, an expression tree would be transformed into a pre-order sequence, in-order sequence and post-order sequence. For each traversal type, we create a dataset and task, where the label is a traversal-specific sequence. This method would give us three times more data for the training purpose, and it could also help the encoder learn semantics in a robust fashion.

\subsection{Shared Encoder Multi Decoder Architecture}

Our encoder-decoder model is based on transformers. It is understood that the encoder learns semantics about the inputs and the decoder attempts to use those semantics with encoder-decoder attention and self-attention to generate the expected sequence. Like bottom layers of the neural network learn syntactically, and micro semantics like POS tagging, dependency relations, etc about the natural language and upper layers closer to the loss function would try to learn higher-order semantics useful for the decision making to minimize loss. Similarly, the encoder would attempt to learn semantics specific to the input, and the decoder will try to leverage it to generate a sequence that minimizes the loss function attached to decoder generated sequence. We use cross-entropy loss for generating a sequence of tokens. 

Since, we have three variants of tasks, MWP-pre-order, MWP-post-order, and MWP-in-order, we have two different design choices available. Either we could append an extra token specific to the task type in the input and use common-encoder common-decoder architecture, or we could choose to not append any task-specific hint and instead use a task-specific decoder to generate the sequence in an expected way. Our proposed model MMTM uses the latter option, with the motivation that the encoder would be able to learn common semantics, i.e. graph and tree relationships, pertaining to the input MWP, and the decoder would be directed to learn the traversal sequence of the tree. We also conduct an analysis to empirically validate this design choice in a later section.

\subsection{Multi-Tasking Pre-Training}

Generally, deep learning models are trained in two stages, pre-training, and fine-tuning. Mostly, pre-training is self-supervised training on a much larger dataset, which would help the model learns broader semantic representation for tokens. An example, BERT uses mask language modelling and next sentence prediction as learning tasks for its pre-training. The fine-tuning stage is more specific to the downstream task. Generally, fine-tuning is performed in a supervised fashion on a dataset which is smaller than the pre-training data corpus. The fine-tuning stage will further contextualize and transform representations of vocabulary tokens to minimize the loss function specific to the downstream task. We introduce a novel pre-training method for training model on MWPs, where we feed our model with variants MWPs tasks under multi-tasking settings, which would help encoder learning semantics of input math word problems in a more robust way. Multi-tasking setting is a process where the same model is given multiple tasks to learn, and it is proven to be more effective in various scenarios.

MMTM model is trained on three tasks MWP-pre-order, MWP-in-order, and MWP-post-order for multiple epochs. After which we drop the decoders specific to the variant tasks and only keep the desired task for the fine-tuning. Learning of shared encoder from multi-tasking would be transferred to the fine-tuning stage.

We also validate the impact of transfer learning, by initializing embeddings for the vocabulary from a pre-trained model, in comparison to the random initialization. We present the empirical results of this study in the later section.

\section{Experiments}

We perform our experiments by training on datasets MWPS and ASDiv-A and test them on the SVAMP dataset, as prescribed by the adversarial SVAMP challenge. MWPS is a widely used dataset composed of 2737 MWPs. ASDiv-A is a subset of ASDiv, which has 1218 MWPs with arithmetic operations. SVAMP challenge dataset is an adversarial dataset to test the model’s generalizability, and compare it with the performance of the SOTA models. SVAMP challenge, research settings, and results of the experiments are explained in the following sections.

\subsection{SVAMP challenge}

SOTA models like the graph-to-tree model, and goal-oriented tree structure model have achieved high accuracy on elementary school level MWPs datasets like MAWPS, MathQA, and Math23K, etc. The focus of researchers is getting shifted to solving complex MWPs on integration, differentiation, etc. However, such SOTA model failed miserably on diverse and adversarial datasets like ASDiv and SVAMP. One of the key reason for models’ high performance on the above datasets were attributed to lower lexical diversity. ASDiv dataset was created by choosing questions to have higher lexical diversity. SVAMP is an adversarial challenge set of 1000 MWPs manually created by experts using nine adversarial variations on 100 MWPs.  The examples in SVAMP test a model across different aspects of mathematical ability to solve word problems. These adversarial variations are of three types, (i) question sensitivity variation, (ii) reasoning ability variation, and (iii) structural invariance variation. Question sensitivity variation would expect the model to be sensitive to the entity on which the question is being asked, reasoning ability variation would test the model to handle variation in semantic mathematical relationships, and structure variation changes the syntactic structure of the question.

\begin{table*}[hbt!]
\begin{tabular}{lcccccccc}
\hline Dataset & Seq2Seq & Seq2Seq & GTS & GTS & Graph2Tree & Graph2Tree & MMTM & MMTM \\
 & (S) & (R) & (S) & (R) & (S) & (R) & (S) & (R) \\
\hline
Full Set & 24.2 & 40.3 & 30.8 & 41 & 36.5 & 43.8 & 47.9 & \textbf{52.3} \\
\hline
One-Op & 25.4 & 42.6 & 31.7 & 44.6 & 42.9 & 51.9 & 48.95 & \textbf{57.08} \\
Two-Op & 20.3 & 33.1 & 27.9 & 29.7 & 16.1 & 17.8 & \textbf{44.72} & 37.13 \\
\hline
ADD & 28.5 & 41.9 & 35.8 & 36.3 & 24.9 & 36.8 & 51.29 & \textbf{68.39} \\
SUB & 22.3 & 35.1 & 26.7 & 36.9 & 41.3 & 41.3 & 42.77 & \textbf{45.77} \\
MUL & 17.9 & 38.7 & 29.2 & 38.7 & 27.4 & 35.8 & \textbf{56.07} & 42.99 \\
DIV & 29.3 & 56.3 & 39.5 & 61.1 & 40.7 & 65.3 & 55.08 & \textbf{60.47} \\
\hline
\end{tabular}
\caption{Comparing Results of MMTM model on SVAMP challenge}
\label{tab:mmtmresults}
\end{table*}

\subsection{MMTM Configuration}

We have experimented MMTM model with varying embedding dimensions from 32 to 768. We have also tested with 1 and 2 layers in encoders and decoder. We initialize MMTM vocabulary with random initialization or from pretrained ROBERTa embeddings. In order to support lower-dimensional MMTM, we have used Principal Component Analysis to identify the most significant dimensions to initialize MMTM vocabulary embeddings. MMTM achieves the best performance with the dimensionality of 62, and a detailed analysis of its performance with respect to change in embedding dimensions is given in the later section. Other hyper-parameters are mentioned in the appendix for reference. Quantities present in MWPs are replaced with placeholders n1, n2, etc in the positional order, which would help to reduce decoder vocabulary significantly, as it does not have to generate actual numbers but just the placeholders. Actual quantities would replace placeholders at the time of expression tree evaluation to compute the final answer. We perform one epoch of pre-training and 3 epochs of fine-tuning. As the dataset size is smaller, the model tends to get overfit if we train for more epochs. The learning rate in pre-training is kept smaller at 10\^-5 than the learning rate in fine-tuning at 10\^-4. We report our accuracy based on the computed answer by evaluating the expression trees and matching them with gold truth answers. 

\subsection{Results}
We compare the results of the MMTM model with the other three SOTA models on the SVAMP challenge set. Three SOTA models are Seq2Seq model, GTS model, and Graph2Tree models. Seq2Seq consists of a Bidirectional LSTM Encoder and an LSTM decoder with attention \cite{luong2015effective} to generate the equation from an input math word problem. GTS \cite{xie2019goal} uses an LSTM Encoder and a tree-based Decoder. Graph2Tree \cite{zhang2020graph} model has a Graph-based Encoder with a Tree-based Decoder. Results of these models on the SVAMP challenge are taken from \cite{patel2021nlp}. 

We report two settings for all these four models, embeddings for their vocabulary are learned from scratch (S), or are initialized from the pre-trained RoBERTa model (R). We report the accuracy of the model on the SVAMP challenge as the performance metric, it is computed by comparing answers after evaluating the predicted expression tree with the ground truth answer. As we can notice in Table \ref{tab:mmtmresults}, the MMTM model significantly outperformed all other SOTA models in both settings. MMTM (R) model report an accuracy of 52.3\% as compared to the 43.8\% performance of the Graph2Tree (R) model, which is a relative improvement of 19.4\%, on the other hand, MMTM (S) achieves the performance accuracy of 47.9\% in comparison of 36.5\% of Graph2Tree (S) model, with the relative improvement of 31.23\%. MMTM model successfully demonstrated how to leverage transformer by training on limited-sized datasets, with robust performance on the adversarial test datasets.

\section{Analysis}

\subsection{Performance on Cohorts of MWPs}

The improvement achieved by the MMTM model is robust across all different cohorts of math word problems, i.e. one operator-based MWPs, and two operator-based MWPs. Similarly on the type of operator based cohorts of MWPs, i.e. ADD, SUB, MUL, and DIV. As we can see in Table \ref{tab:mmtmresults}, the best baseline performance on two-operator MWPs was achieved by GTS (R), which was significantly outperformed by the MMTM model with a relative improvement of 50.1\%. Such a gain in the performance on relatively difficult MWPs demonstrates the better ability of the MMTM model to generalize. Similarly, all the baseline models have poor comparative performance on MUL MWPs, which is not the case with the MMTM model as it reports approximately similar performance across all the types of operator-based MWPs cohorts.

\subsection{Impact of Embeddings Dimensions}

\begin{figure}[h]
    \centering
    \includegraphics[scale=0.575]{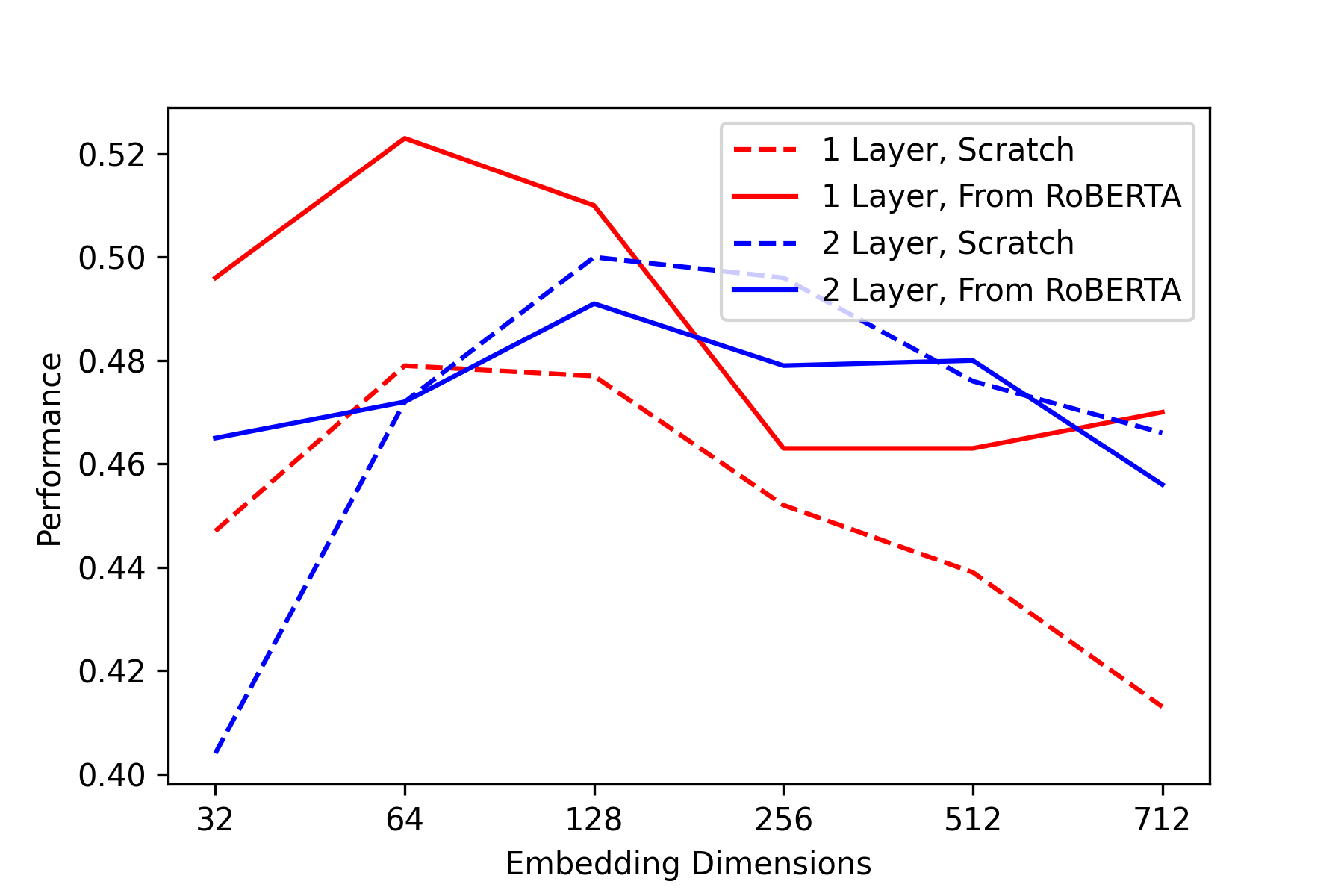}
    \caption{Impact of Embeddings Dimensions}
    \label{fig:math_results}
\end{figure}

Figure \ref{fig:math_results} presents the comparison of MMTM performance when its dimensionality of input vocabulary and hidden representations in transformer layers are varied from 32 to 768. Figure \ref{fig:math_results} compares four different settings, MMTM (S) and MMTM (R) with 1 transformer layer and 2 transformer layers. As we could notice, MMTM gives an optimal performance on current datasets with lesser model complexity, which has lower dimensionality of 62 and single layer transformer. This could be mostly attributed to the limited-sized dataset for the training. For almost all the data points, MMTM with vocabulary initialized from Roberta gives better performance than the MMTM with weights learned from scratch. This demonstrates how to leverage transfer learning for such settings, and it remains more effective at lower dimensions. The performance gap at higher dimensions is narrower for both 1L and 2L settings between MMTM with and without transfer learning.  

\subsection{Ablation Study}

\begin{table}[hbt!]
\fontsize{9pt}{14pt}
\selectfont
\begin{tabular}{lc}
\hline Model Variants & SVAMP Accuracy  \\
\hline
MMTM & 52.30 \\
MMTM (Scratch) & 79.51\\
MMTM (without Multi Tasking - \\\hspace{0.75cm}Multi Decoder Pretraining) & 43.40 \\
MMTM (with 768 Embedding - \\\hspace{0.75cm}   Dimensions) & 47 \\
\hline
\end{tabular}
\caption{Ablation study of MMTM Model}
\label{tab:ablation}
\end{table}

We have made key design choices for our MMTM model, which are mainly (i) pre-training with multi-tasking multi-decoder settings (ii) lower dimensionality of its representations, i.e. 64 (iii) PCA on pre-trained RoBERTa embeddings to initialize input vocabulary embeddings of MMTM. We carried ablation study to know the impact of each such decision and the results could be seen in Table \ref{tab:ablation}. We could observe that Multi-Tasking Multi Decoder based pre-trained is the most effective design choice with an impact of 8.9\% in the accuracy. This reinforces our hypothesis that the shared encoder learns semantics from input MWPs in a generic and robust way, which is transferred to the fine-tuning stage, which achieves better performance on a difficult adversarial challenge set. The default embeddings size of transformer models like BERT, RoBERTa, T5 etc is 768. We have empirically found that the higher embeddings dimension increases the complexity of the model, which tends to underperform on the test set when it is trained on a smaller training corpus. The performance of the MMTM model drops by 5.3\% when we use 768 dimensions for its embeddings.  Initializing vocabulary embeddings from principal components of pre-trained RoBERTa’s embeddings also has an impact of around 4.4\%.

\subsection{Model Interpretability}
\begin{table*}[hbt!]
\fontsize{8pt}{14pt}
\selectfont
%\begin{tabular}{ccccc}
\begin{tabular}{|p{0.5cm}|p{1.5cm}|p{8cm}|p{3.5cm}|p{1.3cm}|}
\hline Class & Model & Question & Predicted Label & Correctness \\
\hline
%-------

\multirow{2}*{RR} & Transformer & \includegraphics[width=8cm]{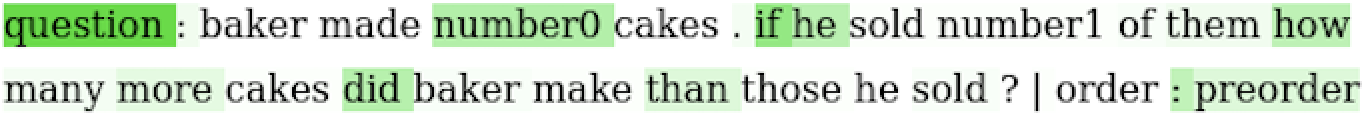} & - number0 number1 & Correct \\
& MMTM & \includegraphics[width=8cm]{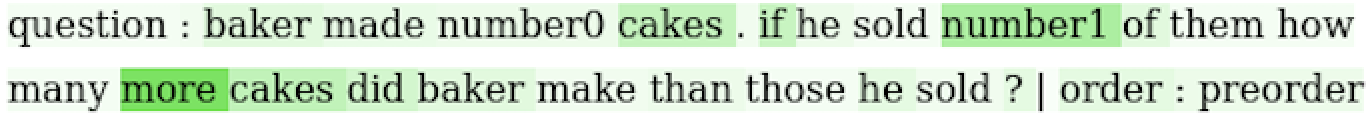} & - number0 number1 & Correct \\
\hline
\multirow{2}*{RR} & Transformer & \includegraphics[width=8cm]{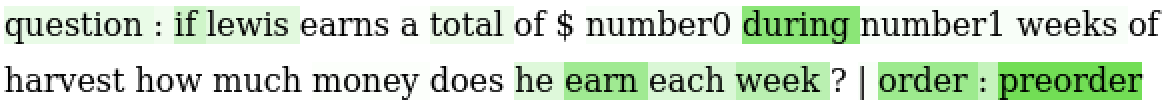} & \slash \hspace{0.05cm}  number0 number1 & Correct \\
 & MMTM & \includegraphics[width=8cm]{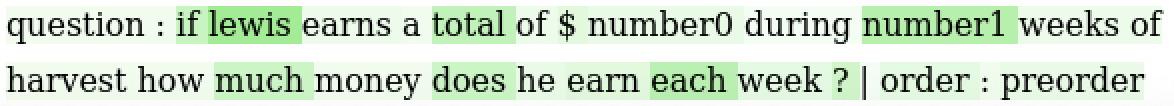} & \slash \hspace{0.05cm}  number0 number1 & Correct \\
\hline
%-------

\multirow{2}*{WR} & Transformer & \includegraphics[width=8cm]{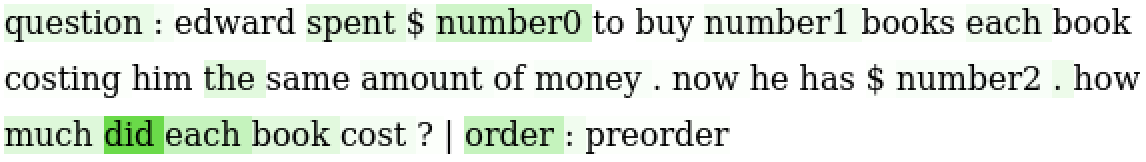} & * number1 number2 & Incorrect \\
 & MMTM & \includegraphics[width=8cm]{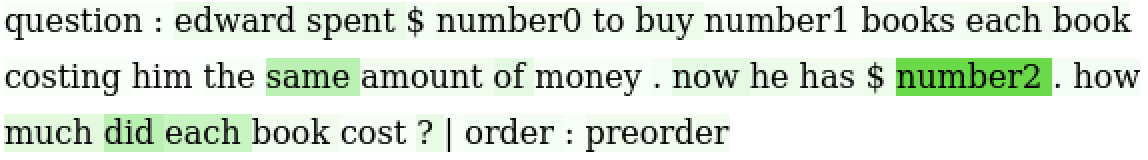} & \slash \hspace{0.05cm} number0 number1 & Correct \\
\hline

\multirow{2}*{WR} & Transformer & \includegraphics[width=8cm]{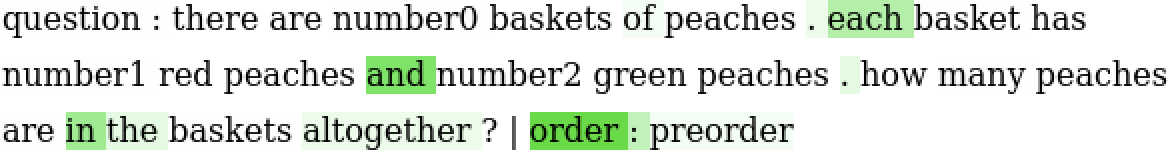} & * number1 number2 & Incorrect \\
 & MMTM & \includegraphics[width=8cm]{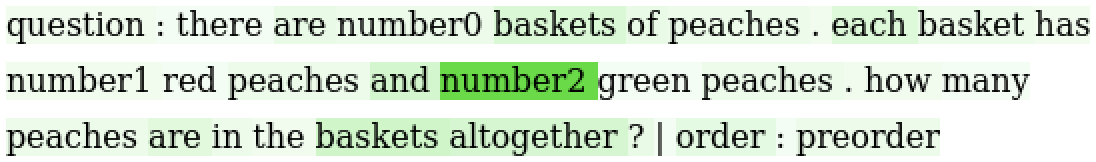} & * number0 + number1 number2 & Correct \\
\hline
%-------

\multirow{2}*{RW} & Transformer & \includegraphics[width=8cm]{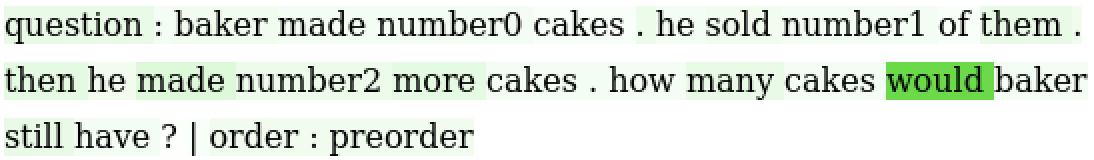} & - + number0 number2 number1 & Correct \\
 & MMTM & \includegraphics[width=8cm]{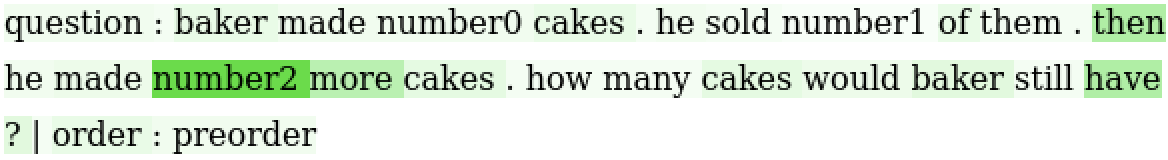} & + number0 number1 & Incorrect \\
\hline

\multirow{2}*{RW} & Transformer & \includegraphics[width=8cm]{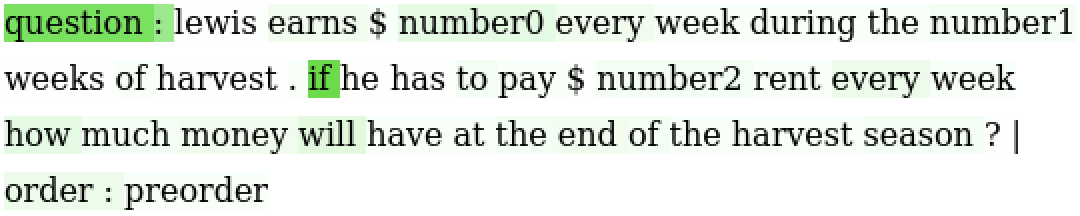} & * number1 - number0 number2 & Correct \\
 & MMTM & \includegraphics[width=8cm]{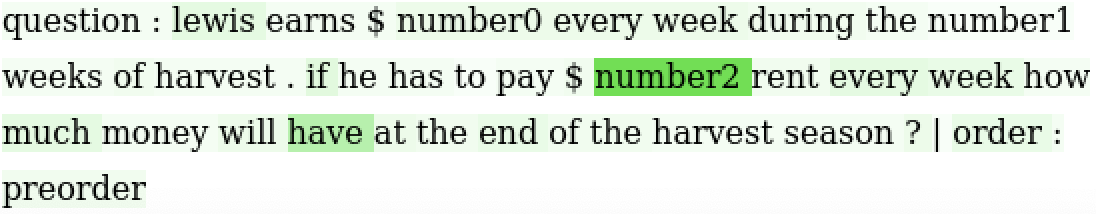} & - + number0 number1 number2 & Incorrect \\
\hline
%-------

\multirow{2}*{WW} & Transformer & \includegraphics[width=8cm]{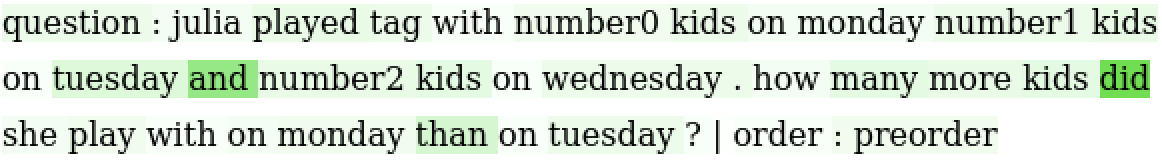} & - + number0 number2 number1 & Incorrect \\
 & MMTM & \includegraphics[width=8cm]{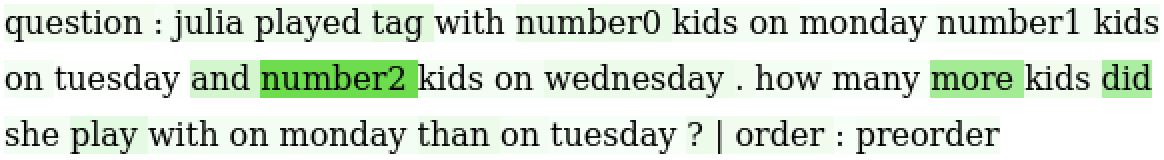} & + number0 number1 & Incorrect \\
\hline
\multirow{2}*{WW} & Transformer & \includegraphics[width=8cm]{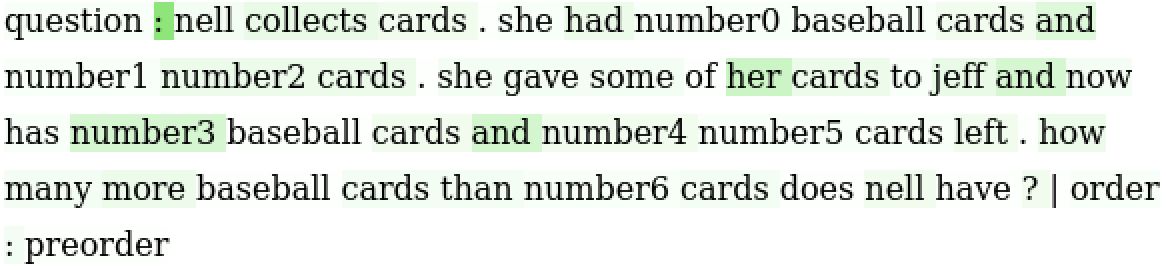} & - + number0 number1 number2 & Incorrect \\
 & MMTM & \includegraphics[width=8cm]{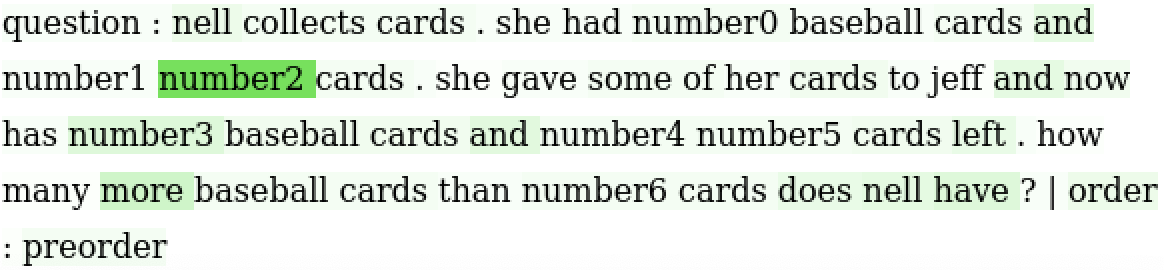} & - + number0 number1 number2 & Incorrect \\
\hline
\end{tabular}
\caption{Attention Analysis: How Transformer and MMTM models attend words from input math word problems.}
\label{tab:expablation}
\end{table*}

There has been increasingly growing attention on interpreting the model and its internal mechanism \cite{gaur2021semantics}. The attention mechanism is at the foundation of transformer architecture \cite{vaswani2017attention}. We have investigated a few examples to study if the MMTM model has a better ability to channelise attention to solve math word problems. We have categorised examples into four classes, (i) Both baseline transformer and MMTM model gave correct prediction (RR), (ii) Baseline transformer model gave the wrong prediction but MMTM model gave correct prediction (WR), (iii) Baseline transformer model gave correct prediction but MMTM model gave wrong prediction (RW), and (iv) Both baseline transformer and MMTM model gave wrong prediction (WW). We took a couple of examples from each class and visualise their attention, which can be seen in Table \ref{tab:expablation}. For the RR class, we can notice baseline transformer model gives strong attention to words like 'question', 'preorder', however, MMTM model has diverse attention with strong attention being given to words like 'number1', 'more', 'total', 'each' etc. For WR class, we can notice baseline transformer has strong attention to some common English words like 'did', 'and' and 'in', whereas the MMTM model attends to mathematically meaningful words like 'same', 'number2', 'each', 'altogether' etc. For the RW class, even though the baseline transformer has given correct prediction, it strongly attended to words which do not make mathematical reasoning sense, as it strongly attends to words like 'question', 'would', 'if', etc. We can also clearly observe the attention pattern difference even in cases where both the models have got its prediction wrong. Attention given by the baseline transformer does not appear to be mathematically reasonable, but MMTM channelises attention more meaningfully, which could be the main reason behind its performance on the adversarial dataset.

\section{Summary and Path Forward}
We presented a novel architecture to solve Math Word Problems. The mathematical reasoning ability of models is important to achieve robust performance and generalize across datasets. There is decent progress to solve elementary level MWPs with datasets like MWPS, AsDiv, Math23K, and MathQA. However these datasets are not bigger enough and models like GTS, Graph2Tree achieved good performance \cite{xie2019goal}, \cite{zhang2020graph}. However when these models are trained over MWPS and AsDiv and tested over an adversarial challenge dataset, SVAMP, their performance drops significantly. This could be mainly because these SOTA models learn shallow heuristics on these limited sized datasets. On the other hand transformer models performance with near-perfect accuracy on very large synthetic datasets on complex, MWPs  \cite{saxton2019analysing} \cite{lample2019deep}. We presented a set of techniques and design choices to leverage transformers on limited sized datasets with much better performance on the adversarial challenge dataset, SVAMP. Multi-Tasking Multi-Decoder model leverage the tree structure of expression trees, which are the output labels for MWPs. We generate three variant tasks with the labels derived using pre-order, in-order and post-order traversal of expression trees. MMTM model training comprises of pre-training stage with multi-tasking and task-specific decoders on these variant tasks. It guides the encoder model with the ability to understand tree semantics of expression trees, and a task-specific decoder could focus on learning traversal order. Our model achieves a relative improvement of 19.4\% on the SVAMP challenge dataset. Model's performance across different cohorts of SVAMP datasets remains robust. We carried out an ablation study to measure the impact of each of our design choices. 

In future, we plan to empirically test if a tree decoder using a transformer could further improve performance and mathematical reasoning ability. Also, we plan to validate if decoupling numerical representation and descriptive representation of quantities present in MWPs as an architectural choice in the encoder is similar to better natural language understanding achieved by DeBERTa model \cite{he2020deberta}. Infusing external knowledge from knowledge graphs to better understand the entities and their relationships of MWPs would be another direction to take the research forward. Overall, solving MWPs reliably could be helpful in building applications to deliver learning outcomes and could help in automated content creation. 

\bibliography{mmtm}
\bibliographystyle{acl_natbib}

\end{document}